\documentclass[a4paper,fleqn]{cas-dc}

\usepackage[numbers]{natbib}
\usepackage{amsmath,amssymb,amsfonts}
\usepackage{enumerate}
\usepackage{booktabs}  
\usepackage{threeparttable}
\usepackage{multirow}
\usepackage{graphicx}
\usepackage{textcomp}
\usepackage{xcolor}
\usepackage{algorithm}
\usepackage{algpseudocode}
\usepackage{float}
\usepackage{lipsum}
\usepackage{CJK}

\def\tsc#1{\csdef{#1}{\textsc{\lowercase{#1}}\xspace}}
\tsc{WGM}
\tsc{QE}
\tsc{EP}
\tsc{PMS}
\tsc{BEC}
\tsc{DE}

\begin{document}
\begin{sloppypar}
\let\WriteBookmarks\relax
\def\floatpagepagefraction{1}
\def\textpagefraction{.001}
\shorttitle{Financial Ticket Faster Recognition}
\shortauthors{F Tian et~al.}

\title [mode = title]{Research on All-content Text Recognition Method for Financial Ticket Image}

\author[1]{Fukang Tian}[style=chinese]
\author[1]{Haiyu Wu}[style=chinese]
\author[1]{Bo Xu}[style=chinese, orcid=0000-0002-7304-857X]
\cormark[1]

\address[1]{Xi'an Network Computing Data Technology Co., Ltd., Xi'an 710049, China}

\cortext[cor1]{Corresponding author}

\begin{abstract}
With the development of the economy, the number of financial tickets increases rapidly. The traditional manual invoice reimbursement and financial accounting system bring more and more burden to financial accountants. Therefore, based on the research and analysis of a large number of real financial ticket data, we designed an accurate and efficient all contents text detection and recognition method based on deep learning. This method has higher recognition accuracy and recall rate and can meet the actual requirements of financial accounting work. In addition, we propose a Financial Ticket Character Recognition Framework (FTCRF). According to the characteristics of Chinese character recognition, this framework contains a two-step information extraction method, which can improve the speed of Chinese character recognition. The experimental results show that the average recognition accuracy of this method is 91.75\% for character sequence and 87\% for the whole ticket. The availability and effectiveness of this method are verified by a commercial application system, which significantly improves the efficiency of the financial accounting system.

\end{abstract}

\begin{keywords}
Ticket detection\sep Image text recognition\sep Financial accounting\sep Deep learning
\end{keywords}

\maketitle

\section{Introduction}
With the rapid development of computer hardware and computer vision, deep learning methods have been widely used in various fields\cite{miikkulainen2019artificial,feng2019computer,charniak2019introduction,solis2019domain}. The financial accounting is one of the important areas\cite{jha2019automation,srivastava2019optical}. Traditionally, financial accounting work needs to be manually done by accountants. First, accountants need to classify different types of financial tickets, such as Value-Added Tax (VAT) invoice (ordinary invoice, electronic invoice, special invoice), bank ticket, toll ticket (highway ticket, vehicle occupancy fee ticket, highway toll ticket, etc.). Then the basic information of these financial tickets is manually entered into the financial software to generate the corresponding types of accounting vouchers, and each financial ticket is sequentially attached to the corresponding categories of accounting vouchers. Finally, the accountant must double-check if there are any missing tickets and if the sequence of tickets is correct. This workflow is obviously inefficient. Due to the large number and variety of financial tickets, accountants' classification workload is large and time-consuming, labor intensity is high, accounting industry labor cost is high,and work efficiency is low. The accuracy of input information is also greatly affected. Therefore, in order to make accounting more accurate, efficient, and highly automated, optical character recognition (OCR) technology has been gradually applied to the field of financial ticket recognition\cite{srivastava2019optical,ha2017recognition,shreya2019optical}.

An intelligent financial ticket recognition system can reduce work tasks and pressure, improve work efficiency, and effectively reduce labor costs. At the same time, it can also promote the digital, structured, and intelligent storage of accounting information, and facilitate the audit of accountants. In this system, the most important task is to detect and recognize the text information in the content of financial tickets. In order to improve the efficiency and accuracy, we divide the financial tickets into fixed form and non-fixed form according to the shape and semantic features of the text area in the image of financial tickets. The fixed form type can be further subdivided into simple vocabulary types and complex vocabulary types. Through the statistical analysis of 716872 financial tickets produced by 276 companies in China in 2019, the fixed form tickets accounted for 68.27\%, and the non-fixed form tickets accounted for 31.73\%. The focus of this paper is the non-fixed form tickets. Due to the uncertainty of the text shape and semantics, all contents detection and recognition are needed, and then the ticket information is given in the form of formattings, such as bank receipts, vouchers, and other tickets. At present, the detection and recognition algorithm based on deep learning is usually divided into two steps: text region detection and text recognition. This approach and can achieve good results. However, in the practical application of financial work, we found that the main problem that restricts the industrial application of most algorithms is that the accuracy is not enough. There are three specific reasons: 1) Multiple links such as text region detection, character recognition, and post-processing are connected in series, resulting in error superposition; 2) The underlying feature distribution of Chinese and English mixed text images has certain particularity, which results in the inaccuracy of some algorithms; and 3) There are some special noises such as complex background, abrasion, fragmentary, wrinkle, character overlap, tilt, occlusion, cutting, uneven illumination and so on, which cause some detection and recognition algorithms cannot achieve stable accuracy.

In view of the above problems, we extracted 482 types of 1.1 million pieces of financial tickets data from the actual operation of the financial accounting SaaS system as the dataset and designed a all contents detection and recognition method of financial tickets. The main contributions are as follows:

\begin{itemize}
    \item We propose a method for detecting and identifying the all contents of financial tickets. Based on Pixel Aggregation Network text detection and the two-stage deep learning model of character recognition, this method has stable and high precision, which makes the detection and recognition of financial tickets more efficient and accurate.
    
    \item We propose an FTCRF framework, which uses the process of character segmentation and character recognition. According to the characteristics of Chinese characters, the text area information extraction method is designed to effectively improve the accuracy and speed of character recognition, and make the all contents detection and recognition of financial tickets more efficient.
\end{itemize}

\section{Related work}
\subsection{Text detection}
In recent years, text region detection based on deep learning has achieved remarkable results. These methods can be roughly divided into two categories: anchor-based methods and anchor-free methods. In these methods, some used deep networks or complex structures to achieve high accuracy, while others used simple structures to keep a good balance between speed and accuracy.

\subsubsection{Anchor-based}
Anchor based text region detection is usually inspired by object detection algorithms, such as Faster RCNN\cite{ren2015faster} and SSD\cite{liu2016ssd}. Textboxes\cite{liao2016textboxes} can process text with extreme aspect ratio by adjusting the scale of anchor and the shape of the convolution kernel in SSD. Textboxes++\cite{liao2018textboxes++} further optimizes the text boundary regression method, which can be used for multi-directional text detection. SSTD\cite{he2017single} implemented an attention mechanism to enhance the features of the text region in the feature map and suppress background information, so as to improve the recall rate of small text. SPCNet\cite{xie2019scene} took text detection as a target segmentation problem, which use Mask RCNN\cite{he2017mask} to detect arbitrary text. This method has achieved remarkable results in several benchmark tests.

\subsubsection{Anchor-free methods}
Anchor free methods usually use full convolutional networks (FCN) to deal with text region detection as a text segmentation problem. \cite{zhang2016multi} used FCNs and detect character candidates to estimate text blocks with MSER. In order to distinguish adjacent text regions effectively, PixelLink\cite{deng2018pixellink} predicts text, non-text, and link at the pixel level, and then uses a post-processing algorithm to obtain text box and eliminate noise. EAST\cite{zhou2017east} and DeepReg\cite{he2017deep} use FCN to generate a multi-scale fusion feature map and then carry out pixel-level text block prediction. PSENet\cite{wang2019shape} uses FCN to predict the multi-scale text region and then uses the progressive scale expansion algorithm to reconstruct the whole text area. In a word, the main characteristics of anchor free methods are the generation of text labels and post-processing algorithms. These algorithms are also widely used in the industry.

\subsection{Text recognition}
There are three types of character recognition methods: character recognition based on a shallow model, character or word recognition method based on deep network, word recognition or character recognition based on the sequence.

\textit{Character recognition method based on shallow model}. After the text lines are cut according to characters, character recognition can be regarded as character classification. The usual way is to extract the description features of characters, and then use the classifier to carry out the results. The commonly used algorithms of character cutting include connected component analysis and projection method. These algorithms can efficiently cut the characters from the more standard text image. However, when the background is complex, the characters are diverse and the noise level is high, the connected component analysis and projection method cannot achieve ideal results. In the text classification, the commonly used character features are local moment, HOG\cite{dalal2005histograms}, SIFT\cite{lowe2004distinctive}, and so on. \cite{shi2013scene} proposed a text representation method based on DPM (deformable part model). DPM can adapt to font changes and is robust to noise, blur, and other factors.

\textit{Character and word recognition method based on deep network}. Using a deep network instead of traditional artificial features can greatly improve the recognition accuracy. One aims to classify characters. After cutting characters, CNN is used to classify characters. Deep model classifiers such as VGG\cite{simonyan2014very}, GoogLeNet\cite{szegedy2015going}, ResNet\cite{he2016deep}, DenseNet\cite{huang2017densely}, and EfficientNet\cite{tan2019efficientnet} can complete this work. Another classifies words. CNN's strong expression ability makes it possible to classify in word level, and the number of categories is generally large, such as 90000 in English. \cite{jaderberg2016reading} used this method to achieve the best results at that time.

\textit{The method of word or character recognition based on sequence}. The performance of character and word recognition methods depends heavily on the accuracy of text segmentation. To solve this problem, the method of character recognition based on sequence came into being. This kind of method is very similar to the speech recognition method, which takes the text line as a whole, does not do segmentation, and directly recognizes the character sequence in batch or increment. This method can make full use of the context of text sequence for disambiguation and avoid the irreversible error caused by character segmentation. There are two main structures in this method: CNN + Bi-LSTM + CTC\cite{shi2016end} and CNN + Bi-LSTM + Seq2Seq\cite{sutskever2014sequence}`. Firstly, CNN is used to learn the relationship between adjacent pixels, and then Bi-directional Long Short-Term Memory (Bi-LSTM) is used to learn long context. Finally, CTC (Connectionist Temporal Classification) and Seq2Seq are used as objective functions to optimize the parameters of the whole network.

\subsection{Financial ticket identification}
With the continuous progress of text area detection and character recognition technology, the recognition method of financial tickets has been developed rapidly. In \cite{palm2017cloudscan}, a ticket automatic recognition and analysis system CloudScan without prior information is implemented by RNN. The average recall rate is 89.1\%. \cite{sun2019template} proposed a method of ticket intelligent recognition based on template matching. The method uses prior information to determine the area to be identified, and the recognition accuracy reaches 95\%. In \cite{yi2019dual}, a medical invoice recognition method based on Gaussian fuzzy and deep learning model is proposed, which can effectively extract medical invoice information. All the above methods use deep learning technologies to improve recognition accuracy, but there are also some problems. On the one hand, the types of training data are not diverse enough, which leads to fewer types of supporting tickets in some methods. On the other hand, some recognition methods do not fully consider the possible noise in the ticket image, and cannot achieve the expected effect when there is more serious noise interference.

\section{Framework and modules}
Usually, a text detection and recognition algorithm contain a text detection model and a character recognition model, then these two models are connected to a data pipeline to form an OCR system in the service implementation stage. However, due to the lack of pertinence to the characteristics of financial ticket data, most of the combination of text detection and recognition models cannot meet the requirements of high precision of financial accounting work. This section will start from the specific characteristics of financial tickets, and propose a all contents detection and recognition method for financial tickets, and a two-stage character recognition framework based on the object detection model.

\begin{figure*}
\begin{center}
\centerline{\includegraphics[width=2\columnwidth]{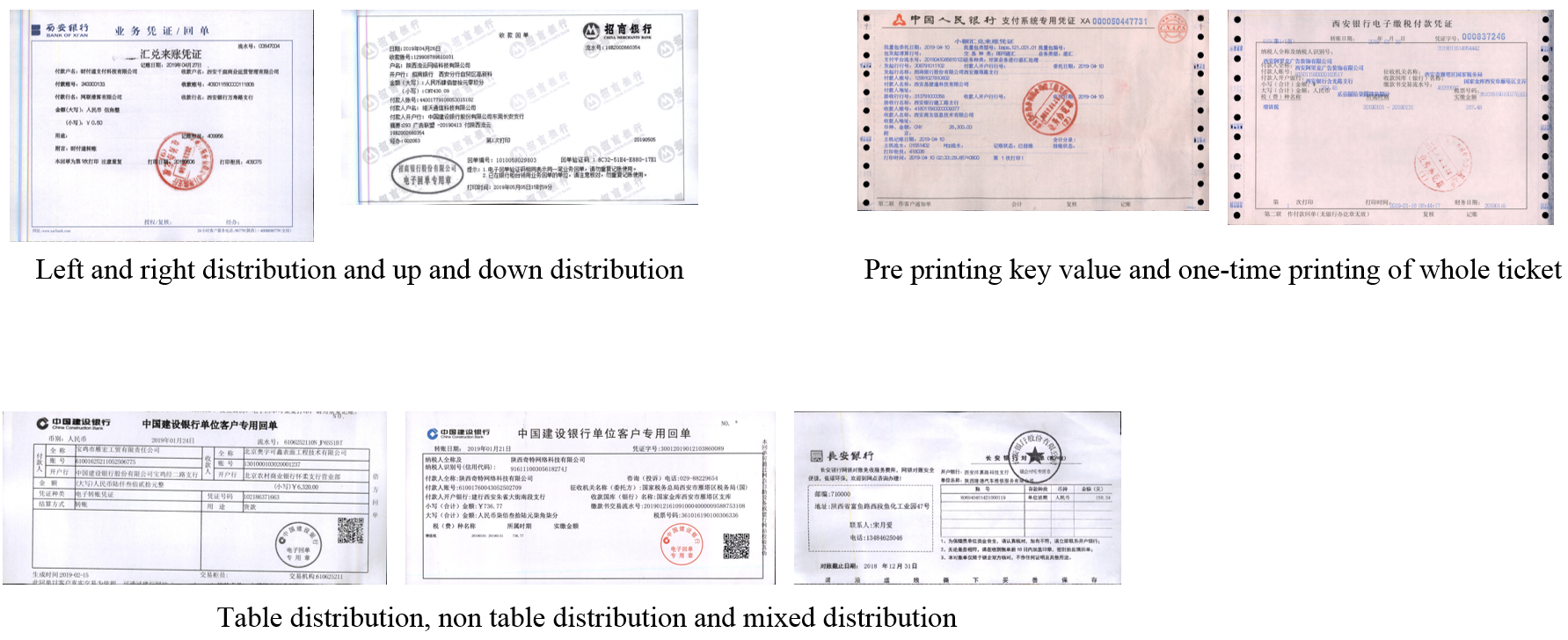}}
\caption{The characteristics of various typesetting of financial tickets.}
\label{fig1}
\end{center}
\end{figure*}

\subsection{Data characteristics}
\begin{figure*}
\begin{center}
\centerline{\includegraphics[width=2\columnwidth]{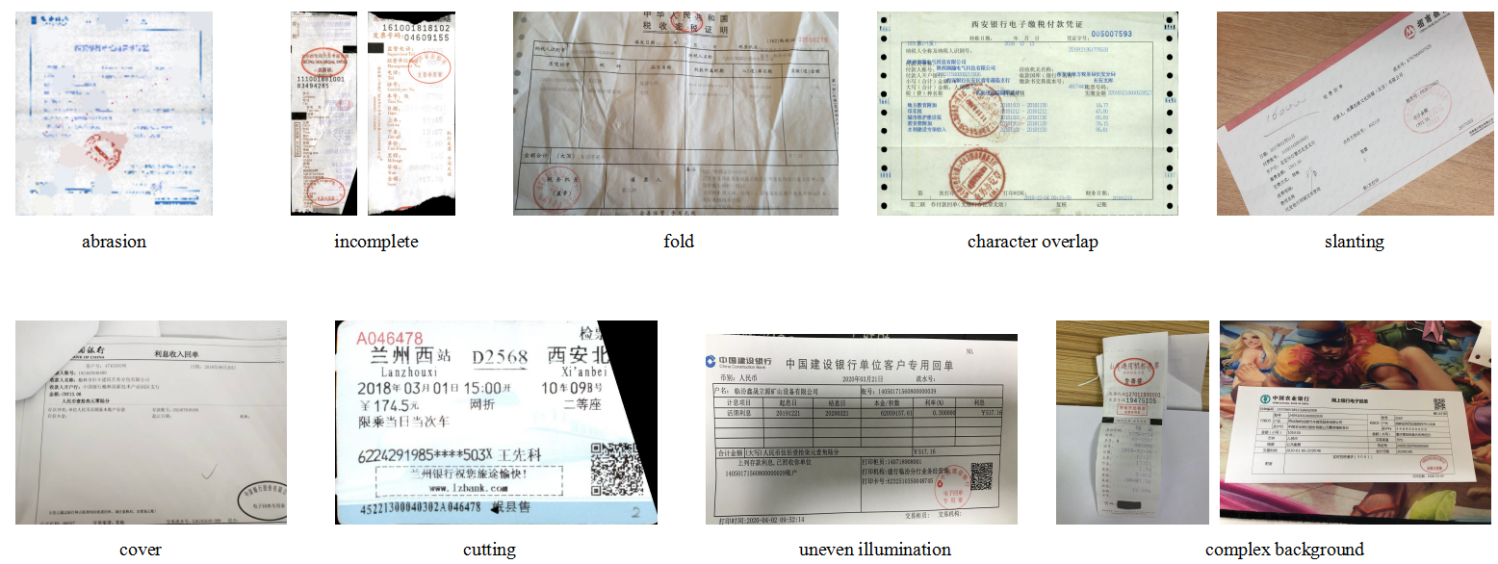}}
\caption{The quality of the ticket itself and the acquisition equipment, technical differences and accidental factors in the collection process will lead to different ticket image quality.}
\label{fig2}
\end{center}
\end{figure*}

Among the 710000 financial tickets collected in this study, 31.73\% of them are non-fixed tickets, with a total of 220000. These tickets come from 182 kinds of receipts, vouchers and detailed tickets from 58 domestic banks in China. Compared with other tickets, the semantic information and text area shape of these tickets are more complex. In financial work, more key information items are required to be extracted. This kind of ticket mainly has the following characteristics:
\begin{enumerate}[1)]
    \item \textit{The types of ticket images are various and the layout is complex}: In China, there are 4034 registered banks. Each bank uses different ticket styles for different business vouchers, and there are tens of thousands of ticket styles. As shown in Figure\ref{fig1}, these ticket styles include left-right distribution, top-down distribution, table distribution, non-table distribution, and mixed table distribution, preprint key values, and one-time printing of the whole ticket. The diversity of ticket style and typesetting makes it more difficult to design a unified and compatible method for text detection and information structure.
    
    \item \textit{The image quality of tickets is uneven}: Financial tickets are usually collected by scanning or photographing. The quality of tickets and accidental factors in the process of the collection will cause noises on the image of financial tickets. As shown in Figure\ref{fig2}, these noises include abrasion, mutilation, character overlap, wrinkles caused by ticket quality, and occlusion, shadow, tilt, and complex background caused by the acquisition process. These noises make detection and recognition more difficult, especially in single character segmentation, the accuracy of traditional algorithms based on manual features is significantly decreased.
    
    \item \textit{There is uncertainty in the distribution of data characteristics}: The uncertainty of financial ticket data characteristic distribution is reflected in three aspects. First, in the actual business, because most banks will regularly optimize their ticket style, or even issue new tickets, so the detection and recognition algorithm will constantly face new ticket types and new template styles. At the same time, the limitations of the training dataset of the detection and recognition model trained by the supervision strategy will make the detection and recognition algorithm appear precision irregularly decline in practical application, which is unable to meet business needs. In addition, new entities of ticket information, such as the newly registered company name or organization name, which does not obey the context sequence feature distribution fitted by the original model in the existing dataset, will make the sequence feature play an opposite role in the recognition process. Lastly, the separability of the mixed data of numbers, Chinese characters, and English letters is relatively reduced, which affects the final character recognition accuracy.
\end{enumerate} 

\subsection{All contents detection and recognition method of financial tickets}

Given the above data characteristics and the low accuracy of traditional ticket recognition methods, we propose a all contents ticket detection and recognition method. This method can adapt to the diversity of ticket types and the uncertainty of its feature distribution, overcome the complex noise, and effectively improve the accuracy of ticket recognition.

The text recognition algorithm based on CTC and attention mechanism models the context information in the text image through the sequence modeling module, which avoids the character segmentation in the text area, and effectively improves the text recognition accuracy in most scenes. However, in the process of financial ticket recognition, we find that the sequence characteristics of key fields such as ticket number, code, amount are not obvious. Most of the new company name or organization name character sequence has greater randomness, which will reduce the recognition accuracy. At the same time, different from English, there are more consecutive overlapping characters of Arabic numerals in financial tickets, which increases the error rate of CTC. This phenomenon also makes the CTC algorithm fail to achieve the expected effect on financial ticket recognition. Therefore, we propose a three-segment character recognition method, which includes three parts: text detection, character segmentation, and character recognition.

\begin{figure}
\begin{center}
\centerline{\includegraphics[width=1\columnwidth]{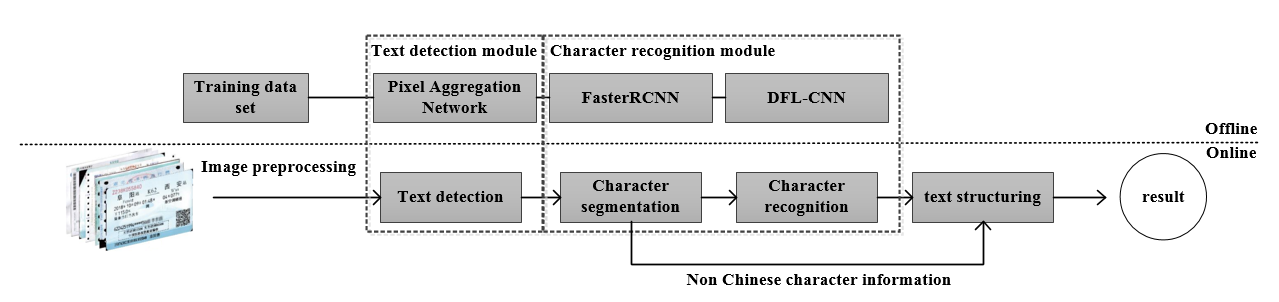}}
\caption{all contents detection and recognition method of financial tickets.}
\label{fig3}
\end{center}
\end{figure}

As shown in Figure\ref{fig3}, the overall method consists of online and offline parts. Each part of the training was supervised offline. Firstly, we manually annotate all text areas in a large number of financial tickets one by one to form a text area detection dataset and use the dataset to train the text area, detection model. Meanwhile, the image of the area to be recognized is cut out one by one from the ticket image, and the single character in the image is annotated manually to form the character segmentation dataset. The object detection model is trained by the dataset to realize the character segmentation. Finally, the character recognition dataset is constructed by the real data and the synthetic data, and the character image classification model is trained to complete the character recognition task. The real data is composed of a single character image cut out from the ticket image. The synthetic data is the character image generated manually according to the texture characteristics of the ticket to solve the problems of unbalanced real data and incomplete covering characters.

In this method, we choose PAN as a text region detection model, Faster RCNN as a character segmentation model, and Discriminative Filter Learning Convolutional Neural Network (DFL-CNN)\cite{wang2018learning} as a character recognition model. After the off-line training, we connect the trained model to the online data stream to form a financial ticket recognition system for all contents recognition. The text area detection model is a text detection module, which is responsible for detecting the location and area of the Chinese text in the ticket. The character segmentation model and classification model constitute the character recognition module, which is responsible for segmenting the single character in the text area and recognizing the semantic information of each character. Finally, according to the keywords of financial work, the identification results with location information are structured, and the key semantic information in tickets is extracted for accounting work. From the daily business, we found that Chinese characters have their unique feature distribution. When English letters and numbers are added to Chinese characters, the separability of data is reduced and the accuracy of character recognition is reduced. Therefore, we adopt a two-step information extraction method, that is to extract non-Chinese character information in the process of character segmentation, so that the classification space of the character image classification model only includes 4103 common Chinese characters, which improves the accuracy of character recognition.

The text region detection of financial tickets needs to have following functions: 1) It can adapt to the complexity of ticket layout, especially in the case of the special interval, table background, and preprint key values dislocation, and still be able to accurately locate keywords and key values, which can improve the accuracy of the final extraction of semantic information; 2) It can better support long text detection; and 3) It can meet the real-time requirements of the financial system and overcome the problem of plenty text areas of bank tickets (the average number of text areas per ticket reaches 28). At present, many deep learning models have excellent performance in text detection. Based on the analysis and comparison of CTPN, EAST, PAN, and other different models, PAN is selected as the text detection model. In the financial ticket scenario, this method has the following advantages:

\textbf{I.} \textit{Arbitrary shape text can be detected}. PAN uses the clustering idea to reconstruct the complete text instance from the kernel. If the text instance is regarded as a clustered group, then the kernel of the text instance is the center of the group, and the text pixels are the clustered samples. Naturally, in order to aggregate text pixels into their corresponding kernels, the distance between text pixels and kernels of the same text should be small enough. Therefore, the loss function is designed as follows:
\begin{equation}
    L = L_{tex} + \alpha L_{ker} + \beta(L_{agg}+L_{dis})
\end{equation}
Where $L_{tex}$ is the loss of the text area, $L_{ker}$ is the loss of the kernel. $\alpha$ and $\beta$ are used to balance $L_{tex}$, $L_{ker}$, $L_{agg}$ and $L_{dis}$. Considering the imbalance between text and non-text pixels, we also use $dice loss$ to monitor the segmentation result $P_{tex}$ and kernel $P_{ker}$, so $L_{tex}$ and $L_{ker}$ is as follows:
\begin{equation}\label{equation1}
    L_{tex} = 1 - \frac{2\sum_{i}P_{tex}(i)G{tex}(i)}{\sum_{i}P_{tex}(i)^{2}+\sum_{i}G{tex}(i)^2}
\end{equation}
\begin{equation}\label{equation2}
    L_{ker} = 1 - \frac{2\sum_{i}P_{ker}(i)G{ker}(i)}{\sum_{i}P_{ker}(i)^{2}+\sum_{i}G{ker}(i)^2}
\end{equation}
Where $P_{tex}$ and $G{tex}$ refers to the $i-th$ result of the segmentation result and the ground truth of the text area, respectively. Similarly, $P_{ker}$ and $G{ker}$ refers to the $i-th$ pixel value of the prediction result and the ground truth of the kernel, respectively. In this way, it can effectively solve the problem of the large length-width ratio of the text area in bank tickets, and improve the accuracy of overlapping text detection to a certain extent.

\textbf{II.} \textit{It is lightweight and real-time}. To improve the efficiency, PAN uses a lightweight segmentation head composed of a feature pyramid enhancement module (FPEM) and feature fusion module (FFM) based on ResNet-18 backbone. This approach can improve the ability of feature extraction and reduce the amount of calculation. Hence, the FPS of the text detection module in the all contents detection of bank tickets can reach 3.67, which meets the real-time requirements of financial work.

After the text region detection is completed, character recognition is also the key to ensure the overall accuracy and speed of this method. Therefore, we propose an FTCRF framework, which is more efficient than traditional algorithms. We will introduce it in detail in the next section. Through the above two modules, in the online part, we will get the recognition results of all strings in the content of a ticket. However, this is not the key information for financial work. Thus, we need to use information structured code to convert strings into key information. So far, we have completed the whole process of the whole content detection and recognition of a financial ticket through this algorithm.

\subsection{Character recognition framework of financial tickets}
\begin{figure*}
\begin{center}
\centerline{\includegraphics[width=2\columnwidth]{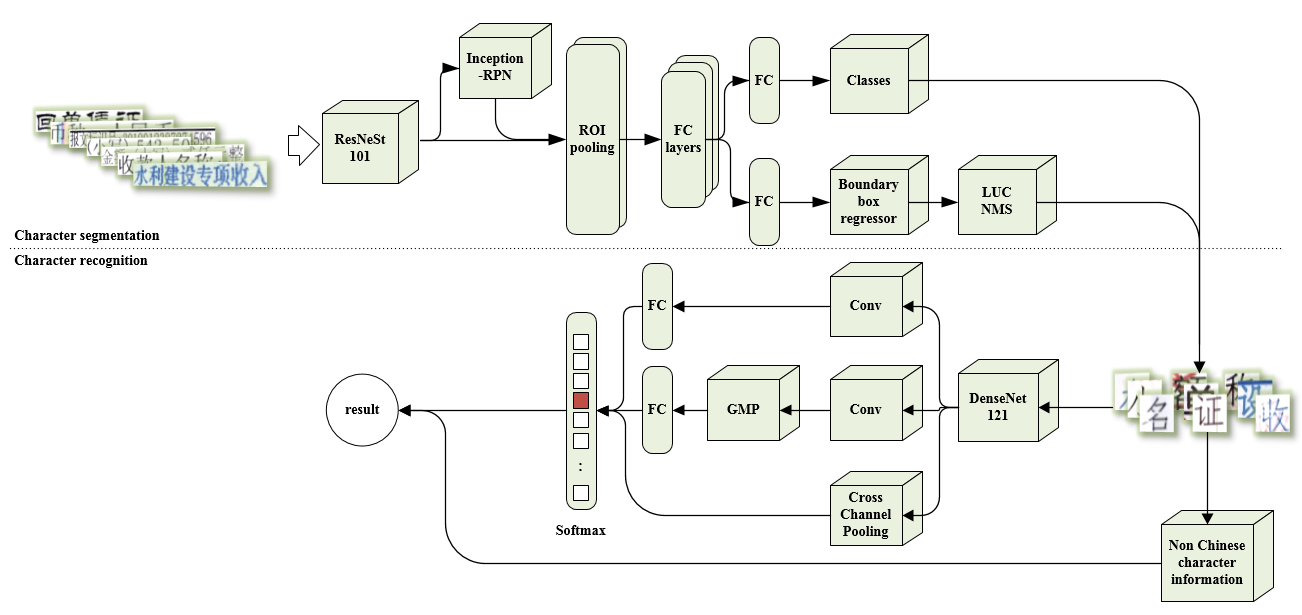}}
\caption{The character segmentation part and character recognition part consist the FTCRF framework.}
\label{fig4}
\end{center}
\end{figure*}
In the process of this method design, we have tried a variety of character recognition algorithms. In addition to the problems existing in the ticket recognition of the end-to-end sequence modeling algorithm mentioned above, we also found that: 1) The accuracy of character segmentation algorithm based on manual features is generally low, and cannot effectively suppress noise; 2) The general convolution neural network model is not sensitive to the Chinese character similar in the form of the image classification task; and 3) Chinese and English mixed information extraction reduces the accuracy of the classifier. To solve these problems, we construct the FTCRF framework based on object detection deep learning model Faster RCNN and fine-grained classification model DFL-CNN. As shown in Figure\ref{fig4}, the network is composed of character segmentation and character recognition. The character segmentation model takes a single character as the target, detects and cuts it out from the text area image, and synchronously returns single-Chinese-character images, and the content and position information of non-Chinese character. The character recognition model uses an image classifier to classify the  received single-Chinese-character images, to obtain the semantic information of Chinese characters in the image, and combine it with the non-Chinese character information provided by the character segmentation model to form the recognition result.

\subsubsection{Character segmentation}
The basic framework of character segmentation is Faster RCNN, which is the result of continuous optimization of regional convolution neural network. It uses the region proposal network and trains it with the whole model. Its objective function includes not only the category and boundary box prediction in object detection but also the binary category and boundary box prediction of anchor frames in the region proposal network. This enables the regional proposal network to learn how to generate high-quality proposal regions, thus ensuring the accuracy of object detection while reducing the number of proposed regions. We use a large number of document text area image data with the character box to train Faster RCNN, and the average precision (AP50) can reach 97\%. For the character segmentation module, the input is the text line image which is cut out after the financial ticket passes the text area detection. The short edge of these images is fixed to 64, and the long edge is scaled according to the scale, then the feature map is extracted from the backbone. On the one hand, local features are extracted through the region proposal network from feature map for text region classification and location regression; On the other hand, combined with the returned location information, unified and standardized region of interest (ROI) regions are obtained through ROI pooling layer. Finally, these ROI regions are classified and further position regression is realized through the classifier and boundary box regression. When we set the detection network category, we divide all Chinese characters into the same category, and English letters, Arabic numerals, and punctuation marks are classified according to their semantic information. Therefore, when the image of the text area passes through the character segmentation network, it outputs the following two contents: 1) The single-Chinese-character image that has been cut out; 2) The semantic information and position information of English letters, Arabic numerals and punctuation marks. The Chinese character image will be sent into the character recognition model, that is, the Chinese character image classifier, to extract the semantic information of Chinese characters. Lastly, all the characters are integrated into strings according to the position information, which is sent to the information structure module as the output of the character recognition module. To further improve the accuracy of Faster RCNN in the ticket character segmentation task, we use the better backbone ResNeSt101, and according to the characteristics of the ticket text region image, we improved the regional proposal network and NMS algorithm.

\begin{figure}
\begin{center}
\centerline{\includegraphics[width=1\columnwidth]{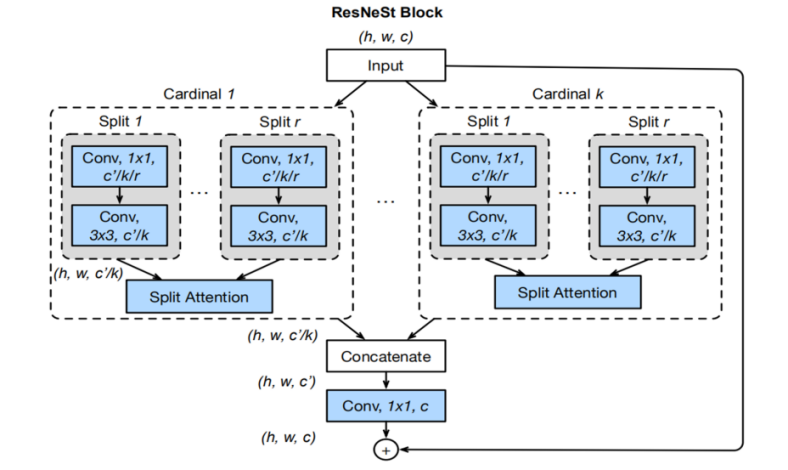}}
\caption{The main structure of ResNeSt block.}
\label{fig5}
\end{center}
\end{figure}

\textbf{Backbone ResNeSt}. The ResNeSt framework is shown in Figure\ref{fig5}. This framework can broaden the network width, realize multi-group feature fusion, and improve the ability of feature extraction of each layer. In the process of feature extraction, the network adopts the idea of GoogLeNet and uses multi-scale convolution. More image features can be extracted by convolution of different sizes. At the same time, all the extracted features are determined by the channel attention, which makes the extracted important features more comprehensive and accurate. Therefore, it is more effective for feature extraction of complex Chinese text. The number of ResNeSt parameters is small, which is helpful to improve the speed of character segmentation. Therefore, we use ResNeSt101 in the character segmentation model.

\begin{figure}
\begin{center}
\centerline{\includegraphics[width=1\columnwidth]{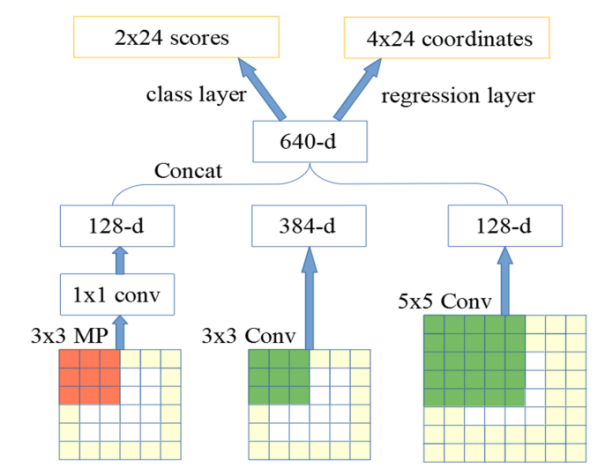}}
\caption{The main structure of Inception-RPN.}
\label{fig6}
\end{center}
\end{figure}

\textbf{Area detection network Inception-RPN}. Most of the character shapes of financial tickets are square areas with an aspect ratio close to 1.0 points, which is a significant feature different from general object detection. In order to accurately detect the character position information, we choose the concept RPN network proposed in deep text network, and its structure is shown in Figure\ref{fig6}. In this paper, nine anchor nodes with three levels and three scales on each pixel are expanded to four levels (4, 8, 16, 32) and 16 anchor nodes (0.8, 1.0, 1.2, 1.5) to make it more suitable for character region detection. At the same time, the network introduces the inception structure of GoogLeNet, which uses multi-scale convolutions, maximum pool, and other methods to extract local features, and forms 640-dimensional feature vectors for text region classification and position regression. Therefore, on the one hand, the network uses multi-scale convolution features, which is more conducive to foreground and background classification. On the other hand, convolution and pooling can effectively extract the regional features of character information.

\begin{figure}
\begin{center}
\centerline{\includegraphics[width=1\columnwidth]{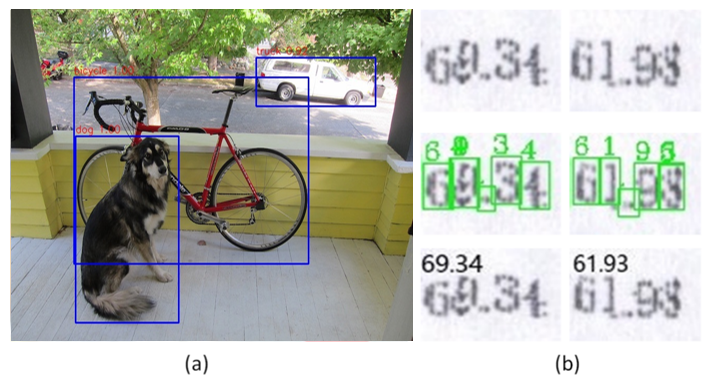}}
\caption{It gives an example for showing the difference between general object detection and text detection.}
\label{fig7}
\end{center}
\end{figure}

\textbf{Location based Unique Character NMS (LUCNMS)}. NMS algorithm is designed for general object detection in Faster RCNN, which supports multi-object overlapping, as shown in Figure\ref{fig7}a. But in financial tickets, character overlap is usually not allowed. If character overlap occurs due to printing problems, only one character can be recognized in the overlapping area, and the remaining characters must be ignored as background. Therefore, the use of a general NMS algorithm may produce multiple character detection results in one location. As shown in Figure\ref{fig7}b, the top second results will lead to the wrong result of financial ticket recognition. Therefore, we propose LUCNMS algorithm for financial ticket recognition, which suppresses the maximum value of all candidate boxes in the same position, so as to improve the accuracy of ticket recognition and greatly improve the speed of candidate box screening.

\subsubsection{Chinese character recognition}
\begin{figure}
\begin{center}
\centerline{\includegraphics[width=1\columnwidth]{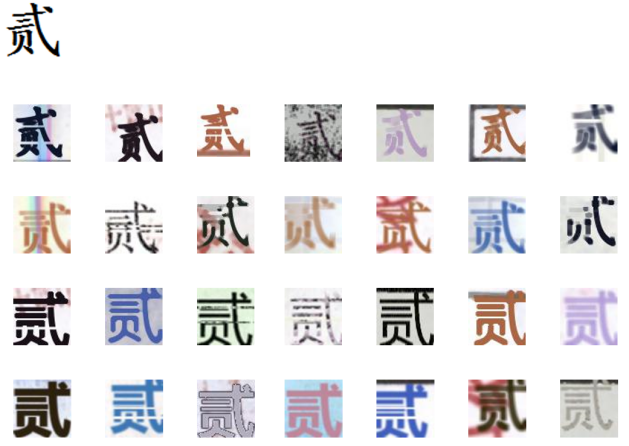}}
\caption{Chinese character synthetic data.}
\label{fig8}
\end{center}
\end{figure}

We use DFL-CNN as a Chinese character recognition network. This network is an end-to-end fine-grained classification network. It uses asymmetric and multi branch structure, completes the extraction of the maximum response region of image classification by combining the $1 \times 1$ convolution layer and the global maximum pooling layer, and strengthens the role of the regional features, so as to achieve the purpose of image classification through subtle differences. It is very helpful to classify the similar characters with slight differences which exist only for the Chinese characters similar in the form. As shown in figure\ref{fig4}, a single-Chinese-character image first forms a feature map through the backbone composed of the first two dense blocks of DenseNet121, and then completes the classification feature extraction through three branches: upper (G-stream), middle (P-stream), and lower (Side branch). G-stream branch is composed of $3 \times 3$ convolution layer and full connection layer, which is used to extract global features. The P-stream branch uses a $1 \times 1$ convolution layer followed by a global maximum pooling (GMP) structure. The maximum response region detected by each convolution kernel is extracted to form key region features (fine-grained features). At the same time, through the Side branch, the Cross-Channel Pooling layer is used to supervise P-stream to extract features around the maximum response region. We choose DFL-CNN fine-grained classifier as the Chinese character recognition network, and improve the accuracy of Chinese character recognition through the following two tasks:

\textbf{Construct Chinese character dataset}. The dataset used for training the DFL-CNN classifier consists of real data and synthetic data. In order to make the classifier have enough generalization ability. Moreover, the training data needs to have diversity, and the feature distribution of the dataset should be consistent with the financial ticket image feature distribution as far as possible. Therefore, as shown in Figure\ref{fig8}, we extract 482 kinds of ticket background texture, character color, font, and other characteristics. A 7-D data construction space is formed by using 4088 common Chinese characters of financial tickets to generate 6.77 million pieces of data, which constitutes the training dataset of the character recognition model.

\begin{figure}
\begin{center}
\centerline{\includegraphics[width=1\columnwidth]{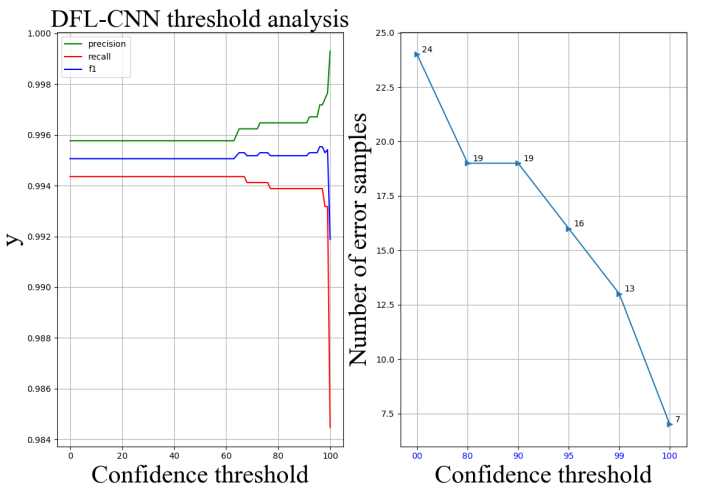}}
\caption{The confidence, recall rate, accuracy, F1 and the number of error classification samples of the DFL classifier are related.}
\label{fig9}
\end{center}
\end{figure}

\textbf{Precision first strategy is adopted}. Because the financial work has high requirements on the accuracy of ticket information extraction, we set the classification confidence threshold and sacrifice the recall rate to further improve the classification accuracy of the model. For the classification results lower than the threshold value, the error correction is carried out by manual correction. Figure\ref{fig9} shows the changes in recall rate, precision, and the number of misidentified Chinese characters with the increase of confidence threshold. From it, when the recall rate of the CNN model is 99.0\% and the threshold value is 99.99\%, it can be seen that the recognition accuracy of the CNN model can reach 99.99\% when the confidence rate of the recall rate is 99.9\%. This feature of high confidence makes DFL-CNN more suitable for the task of financial ticket recognition.


\section{Experiment}
In order to accurately grasp the recognition accuracy, operation speed and operating conditions of the all contents detection and recognition method of financial tickets proposed in this paper, a variety of bank tickets are used to construct datasets, and the functional models in the method are tested.

\subsection{Dataset}
In practical application, we use 74000 bank tickets as the training set to complete the model training. At the same time, as shown in Table\ref{table1}, we collected a total of 2000 tickets from 10 types of banks as the experimental dataset, which was used to complete the model test tasks in this experiment. The dataset contains 28.9 text areas and 339.4 characters per ticket on average.

\begin{table*}
\caption{It shows the supported types of bank tickets and the detailed information for each type.}
\begin{center}
    \begin{tabular}{cccc}
    \hline
    Types of tickets & Average resolution & Average number of ROI & Average number of characters\\
    \hline
    BOC domestic payment receipt & 2456×1720 & 34 & 519\\
    CCB unit customer special receipt & 2480×1164 & 25 & 335\\
    ICBC receipt  & 2464×1174 & 33 & 347\\
    BCM receipt  & 2488×1612 & 31 & 296\\
    CMB receipt  & 2472×1196 & 25 & 284\\
    ABC receipt  & 1656×1160 & 31 & 318\\
    ABC notice  & 2504×1136 & 29 & 376\\
    ABC electronic receipt of online banking  & 1946×1033 & 26 & 297\\
    SPDB receipt  & 2464×1752 & 30 & 336\\
    Bank of Xi'an receipt  & 2366×1810 & 25 & 286\\
    \hline
    \end{tabular}
\end{center}
\label{table1}
\end{table*}

\subsection{Operating environment}
The financial ticket intelligent recognition system is built by Tensorflow and Pytorch framework. The operating system of the system deployment server is CentOS Linux release (Core) 7. And two 32G Tesla V100 graphics cards are used. The main frequency of the CPU is 2.20GHz, and the memory is 256GB.

\subsection{Experimental results}

\subsubsection{Text detection}
In the text region detection experiment, 2000 bank tickets were used as experimental samples to detect 57800 text regions. The experiment compared the current deep learning models of text detection, including CTPN, EAST, PAN512, and PAN640. The experimental results are shown in Table\ref{table2}. From the results, PAN640 has certain advantages in precision and speed, and is more suitable for the text area detection task of financial tickets.
\begin{table}
\caption{The comparison results of CTPN, EAST, PAN512, and PAN640 with the metrics precision (P), recall rate (R), F1, and FPS.}
\begin{center}
    \begin{tabular}{ccccc}
    \hline
    Method & P & R & F1 & FPS\\
    \hline
    CTPN  & 0.83 & 0.84 & 0.83 & 1.10\\
    EAST  & 0.92 & 0.84 & 0.87 & 2.48\\
    PAN512  & 0.89 & 0.92 & 0.90 & 3.67\\
    PAB640  & 0.92 & 0.93 & 0.92 & 2.56\\
    \hline
    \end{tabular}
\end{center}
\label{table2}
\end{table}

\subsubsection{Text recognition}

\begin{figure}
\begin{center}
\centerline{\includegraphics[width=1\columnwidth]{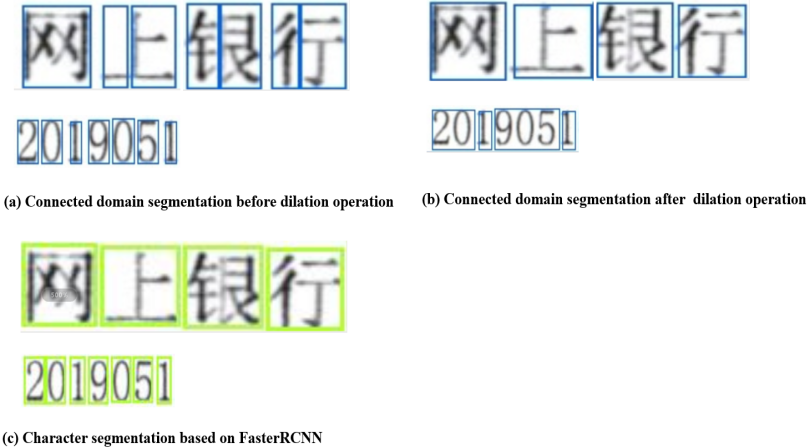}}
\caption{The problems of character segmentation based on connected component analysis algorithm and the example of character segmentation based on Faster RCNN algorithm.}
\label{fig10}
\end{center}
\end{figure}

In the character segmentation experiment, 57800 text region images from 2000 bank tickets were used as experimental data to segment 678800 characters. We compared the traditional character segmentation algorithm based on connected components analysis and the projection method, and compared the effect of Faster RCNN and improved Faster RCNN. Four metrics are used to evaluate the algorithm, which are precision, recall rate, F1 value, and FPS. The results are shown in Table\ref{table3}. It can be seen from the results that the character segmentation algorithm based on connected components analysis is the fastest, but its accuracy is far from meeting the needs of practical application. The characteristics of Chinese characters and complex noise are the main reasons for the errors of connected components analysis and projection method. As shown in figure\ref{fig10}(a), the abrasion of tickets leads to the incoherence of strokes of the character "\begin{CJK}{UTF8}{gbsn}
上
\end{CJK}", resulting in wrong segmentation, and the space between the sides leads to the wrong segmentation of the word "\begin{CJK}{UTF8}{gbsn}
银行
\end{CJK}". If the expansion operation is used to solve such problems, as shown in Figure\ref{fig10}(b), the numbers with close spacing will be adhered to and cannot be correctly segmented. This contradiction is not easy to be solved completely because of the uncertainty of the character width in the financial instruments mixed with Chinese characters, English letters, and numbers. As shown in Figure\ref{fig10}(c), when the target detection model based on deep learning is used to segment characters, it can achieve high precision, and the recall rate is also significantly improved, which is mainly due to the better noise suppression ability of the deep learning algorithm. In the results, the FPS value shows the running speed of each algorithm. In practical application, parallel processing can be used to improve the processing speed by an order of magnitude.

\begin{table}
\caption{The comparison results of Connected components, Projection, Faster RCNN, and ours with the metrics precision (P), recall rate (R), F1, and FPS.}
\begin{center}
    \begin{tabular}{ccccc}
    \hline
    Method & P & R & F1 & FPS\\
    \hline
    Connected components  & 0.43 & 0.73 & 0.54 & 769\\
    Projection  & 0.73 & 0.80 & 0.76 & 31\\
    Faster RCNN  & 0.97 & 0.93 & 0.95 & 20\\
    Ours  & 0.92 & 0.99 & 0.96 & 21\\
    \hline
    \end{tabular}
\end{center}
\label{table3}
\end{table}

Chinese character recognition. Here, the performance of four depth model classifiers, DenseNet-121, ResNeSt-101, EfficientNet-b7, and DFL-CNN in Chinese character recognition tasks are compared. The specific experimental results are shown in Table\ref{table4}. From the results, because DFL-CNN pays more attention to the details of the image and is sensitive to the subtle differences of Chinese characters, it has obvious advantages in the recognition accuracy. Similarly, in practical application, because the resolution of the Chinese character image is smaller than that of the ticket image, parallel processing can be used to improve the processing speed by two orders of magnitude.

\begin{table}
\caption{The comparison results of DenseNet-121, ResNeSt-101, EfficientNet-b7, and DFL-CNN with the metrics precision (P), recall rate (R), F1, and FPS.}
\begin{center}
    \begin{tabular}{ccccc}
    \hline
    Backbone & P & R & F1 & FPS\\
    \hline
    DenseNet-121  & 0.9802 & 0.9811 & 0.9806 & 36\\
    ResNeSt-101  & 0.9860 & 0.9865 & 0.9862 & 34\\
    EfficientNet-b7  & 0.9837 & 0.9841 & 0.9840 & 26\\
    DFL-CNN  & 0.9936 & 0.9912 & 0.9924 & 33\\
    \hline
    \end{tabular}
\end{center}
\label{table4}
\end{table}

\subsubsection{Overall method measurement}
The overall method is measured by string area recognition accuracy and whole ticket recognition accuracy. The string recognition accuracy is expressed by $P_{char}$, which is the proportion of the total number of strings with correct character recognition. The calculation formula is as follows:
\begin{equation}
    P_{char} = \frac{\sum R_{char}}{\sum N_{char}}
\end{equation}
Where $R_{char}$ is the correctly recognized string in a single sample, and $N_{char}$ is the total number of strings in a single sample. The whole ticket identification accuracy is expressed by $P_{ticket}$, which is the proportion of the total amount of financial tickets whose business information fields are correctly identified. The calculation formula is as follows:
\begin{equation}
    P_{ticket} = \frac{\sum R_{ticket}}{\sum N_{ticket}}
\end{equation}
Where $R_{ticket}$ is the total number of financial tickets whose business information fields have been correctly identified, and $N_{ticket}$ is the total number of financial tickets.

The method in this paper consists of three deep learning models, a total of memory occupation is 658M. We use 2000 test data to test the maximum and minimum memory (MaxM and MinM) consumption and recognition speed of this method under the condition of 20 bath size parallel processing. The results are shown in the Table\ref{table5}.

\begin{table*}
\caption{The performance of our method tested on 2000 pieces data.}
\begin{center}
    \begin{tabular}{ccccccc}
    \hline
    Method & $P_{char}$ & $P_{ticket}$ & Batch size & MaxM & MinM & Time(ms)\\
    \hline
    PAN + FTCRF & 91.75\% & 87\% & 20 & 9.5G & 5.2G & 578.33\\
    \hline
    \end{tabular}
\end{center}
\label{table5}
\end{table*}

At present, some studies can achieve high accuracy for specific types of tickets. For example, \cite{liu2020end} uses SSD and CNN-GRU\cite{chung2014empirical} to detect and identify taxi receipts, and the accuracy can reach 94.36\%. However, these studies do not provide better solutions for ticket processing problems that need all contents recognition. As shown in Table\ref{table6}, we compared with some existing methods. From the results, we can see that while supporting more ticket types and all contents detection and recognition, the accuracy can meet the practical application requirements of financial work.

\begin{table}
\caption{The comparison results among our method and some existing methods.}
\begin{center}
    \begin{tabular}{cccc}
    \hline
    Method & all contents & Types & $P_{ticket}$\\
    \hline
    Zhang & - & VAT & 96.21\%\\
    Wu & - & VAT & 95.10\%\\
    Liu & - & Taxi & 94.36\%\\
    Wang & \checkmark & Compound type & 72.98\%\\
    Chen & \checkmark & Compound type & 78.14\%\\
    Ours & \checkmark & Compound type & 87.00\%\\
    \hline
    \end{tabular}
\end{center}
\label{table6}
\end{table}

\section{Conclusion}
In this paper, the all contents detection and recognition method of financial tickets based on PAN320 and FTCRF framework is proposed. It can adapt to the characteristics of financial tickets, support multiple types of tickets and all contents detection and recognition, suppress all kinds of noise. As for the performance, the string recognition accuracy can reach 91.75\% and the single ticket processing time is 578.33ms, which can meet the accuracy and speed requirements of financial work. In addition, the FTCRF framework is proposed, which uses improved Faster RCNN to segment characters to reach higher accuracy and recall rate, where non-Chinese character information is extracted synchronously with character segmentation, and Chinese character information is extracted by DFL-CNN fine-grained classifier. This framework is more suitable for financial ticket recognition task, with higher precision and efficiency.

\bibliographystyle{unsrt}
\bibliography{cas-refs}


\end{sloppypar}
\end{document}